\documentclass{article}

\PassOptionsToPackage{numbers,compress}{natbib}
\usepackage[preprint]{neurips_2026}

\usepackage[utf8]{inputenc}
\usepackage[T1]{fontenc}
\usepackage{hyperref}
\usepackage{url}
\usepackage{booktabs}
\usepackage{amsfonts}
\usepackage{amssymb}
\usepackage{amsmath}
\usepackage{nicefrac}
\usepackage{microtype}
\usepackage{xcolor}
\usepackage{graphicx}
\usepackage{subcaption}
\usepackage{float}
\usepackage{algorithm}
\usepackage{algpseudocode}
\usepackage{multirow}
\usepackage{wrapfig}
\usepackage{enumitem}
\usepackage{placeins}

\DeclareMathOperator{\KL}{KL}

\DeclareMathOperator{\Beta}{Beta}

\newcommand{\methodname}{ATESD\xspace}
\newcommand{\seedstd}[1]{$_{\scriptstyle\pm #1}$}

\newcommand{\tpolicy}{p_{\mathrm{T}}}
\newcommand{\spolicy}{p_{\mathrm{S}}}
\newcommand{\refsol}{y^{\star}}
\newcommand{\rollout}{\hat{y}}
\newcommand{\datacal}{\mathcal{S}}
\newcommand{\loss}{\mathcal{L}}

\usepackage{xspace}

\title{Adaptive Teacher Exposure for Self-Distillation\\in LLM Reasoning}

\author{%
  Zihao Han,
  Tiangang Zhang\thanks{Corresponding author.},
  Huaibin Wang,
  Yilun Sun \\
  ByteDance Douyin \\
  \{hanzihao.3344, zhangtiangang.0909, wanghuaibin, sunyilun\}@bytedance.com
}

\newcommand{\arxivbodyonly}{}

\begin{document}

\maketitle

\begin{abstract}
On-policy self-distillation has become a strong recipe for LLM reasoning, where a privileged teacher supervises the student's own rollouts while conditioning on the reference solution.
A design choice shared by nearly all such methods, however, has gone unquestioned: the teacher always sees the \emph{full} reference reasoning.
We argue that this default itself is part of the problem and identify a \emph{teacher-side exposure mismatch}: when the teacher conditions on reasoning far beyond the student's current competence, the resulting token targets become too strong to absorb.
A controlled fixed-exposure sweep makes this concrete on two fronts: 1) full exposure is not reliably the best choice, and 2) student--teacher mismatch grows monotonically as the teacher sees more privileged reasoning.
This motivates treating teacher exposure not as a fixed hyperparameter but as a learnable training-time control variable.
We therefore propose \textbf{A}daptive \textbf{T}eacher \textbf{E}xposure for \textbf{S}elf-\textbf{D}istillation (\textbf{\methodname}).
\methodname{} models the reveal ratio with a lightweight Beta-policy controller conditioned on compact training-state statistics, and uses one sampled exposure for a short hold window of student updates.
To make this exposure controller learnable, we optimize it with a discounted learning-progress reward that scores each held decision by its effect on the student's \emph{future} improvement rather than its immediate loss change, addressing the delayed credit assignment induced by on-policy distillation.
Experiments on AIME\,24, AIME\,25, and HMMT\,25 across Qwen3-\{1.7B, 4B, 8B\} show that \methodname{} consistently outperforms competitive self-distillation and RL baselines, improving over OPSD by $+0.95$, $+2.05$, and $+2.33$ Average@12 points respectively, and establishing adaptive teacher exposure as an effective new axis for reasoning self-distillation.
\end{abstract}

\section{Introduction}
\label{sec:intro}

Post-training has become the primary route for improving LLM reasoning, with recent progress driven by both reinforcement learning with verifiable rewards~\citep{shao2024deepseekmath,guo2025deepseekr1,yu2025dapo} and distillation-based learning~\citep{hinton2015distilling,agarwal2024onpolicy,gu2024minillm}. Within the latter line, On-Policy Self-Distillation (OPSD)~\citep{zhao2026opsd} has emerged as a particularly clean formulation: a single model plays both teacher and student, the student learns from its own rollouts, and the teacher conditions on a privileged reference solution when providing token-level supervision. By aligning supervision with the trajectories the student actually visits, OPSD removes the \emph{student-side} distribution mismatch that has long limited self-distillation for reasoning, which makes on-policy distillation one of the strongest current recipes for post-training reasoners across model families and scales, and the default backbone of privileged self-distillation pipelines whenever reliable process-level verifiers remain prohibitively expensive to construct. On competition-level mathematical reasoning it is now the dominant route for lifting small open-weight reasoners onto the same accuracy frontier as much larger proprietary teachers from frontier labs.

\begin{figure}[t]
\centering
\includegraphics[width=\linewidth]{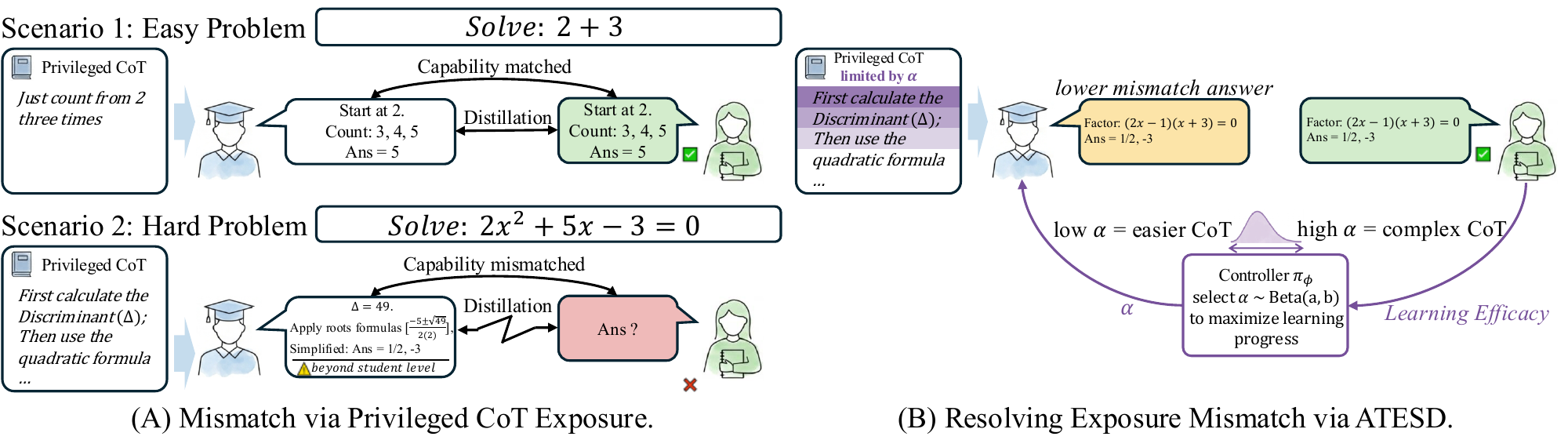}
\caption{Overview of \methodname. \textbf{(A)}~Teacher-side exposure mismatch: on an easy problem (e.g.\ $2{+}3$) the teacher's privileged CoT stays within the student's capability and distillation succeeds; on a hard problem (e.g.\ a quadratic equation) the full CoT far exceeds the student's level, producing targets the student cannot absorb. \textbf{(B)}~\methodname{} limits the privileged CoT via a learned exposure $\alpha$: a Beta-policy controller $\pi_\phi$ selects $\alpha$ to keep supervision absorbable, trained via REINFORCE from a learning-efficacy reward that scores each decision by its effect on future learning progress.}
\label{fig:overview}
\end{figure}

Yet OPSD and its follow-ups fix the student-side mismatch while leaving the teacher side unexamined: \emph{how much} privileged reasoning the teacher itself should see. Existing methods universally adopt full exposure --- the teacher receives the complete reference solution, implicitly treating more information as better supervision. We argue this default is part of the problem and identify a \emph{teacher-side exposure mismatch} (Figure~\ref{fig:overview}A): on easy problems the teacher's reasoning stays within the student's capability and distillation succeeds, yet on hard problems the full privileged Chain-of-Thought far exceeds the student's current competence, producing targets the student cannot absorb. This is the supervision-side analogue of the rollout mismatch that OPSD was designed to remove; the present paper turns exposure into a controllable, learnable variable instead of a fixed assumption during training.

A controlled fixed-exposure sweep (\S\ref{sec:closer_look}) reveals two consistent patterns. First, \emph{Suboptimality of Full Exposure}: intermediate exposure ($\alpha^{*}{=}0.5$) consistently outperforms the full-exposure default across seeds. Second, \emph{Monotonic Mismatch Growth}: teacher--student mismatch grows monotonically with $\alpha$. A coarse difficulty-binned analysis further shows that different learning regimes prefer different tested exposures, suggesting that teacher exposure should be learned from training feedback rather than being fixed to a single universal default across problems and training stages in practice.

Turning exposure into a learnable control variable, however, introduces a training problem: exposure choices affect the student only after subsequent optimization steps. Choices most beneficial to the student's \emph{future} learning often do not yield the largest immediate KD-loss drop, and high-exposure decisions can look unattractive to a single-step proxy. Naive one-step rewards therefore miscredit good exposure decisions, so the controller must instead be trained from delayed learning effects rather than myopic one-step proxy rewards across subsequent student optimization updates.

We address teacher-side exposure mismatch with \textbf{\methodname} (\textbf{A}daptive \textbf{T}eacher \textbf{E}xposure for \textbf{S}elf-\textbf{D}istillation). \methodname{} models exposure as a continuous variable $\alpha \in [0,1]$ parameterised by a lightweight Beta-policy controller over compact training-state statistics. Concretely, the controller selects one global exposure for each hold window and is trained on a two-timescale hold/lookahead schedule: the student updates at every distillation step, while the controller updates more slowly via REINFORCE~\citep{williams1992reinforce} from a discounted learning-progress reward that scores each held decision by its effect on the student's future improvement over subsequent optimization updates during training.

In summary, Our main contributions are three-fold as follows:
\begin{itemize}[leftmargin=*,topsep=1pt,itemsep=0pt,parsep=0pt,partopsep=0pt]
    \item We identify \emph{teacher-side exposure mismatch}, and provide controlled evidence for two patterns --- \emph{Suboptimality of Full Exposure} and \emph{Monotonic Mismatch Growth} --- showing that full exposure is neither the strongest default nor stable as $\alpha$ grows, thereby motivating adaptive control.
    \item We propose \methodname, which treats teacher exposure as a learnable training-state-conditioned variable via a lightweight Beta-policy controller, trained on a two-timescale schedule with a discounted learning-progress reward for held exposure decisions during on-policy distillation training.
    \item On AIME\,2024, AIME\,2025, and HMMT\,2025 with Qwen3-1.7B, 4B, and 8B, \methodname{} consistently outperforms self-distillation and RL baselines, reaches 65.65 Average@12 on Qwen3-4B, and establishes adaptive teacher exposure as an effective new axis for reasoning self-distillation.
\end{itemize}

\section{Related Work}
\label{sec:related}

\subsection{On-Policy Self-Distillation and Teacher--Student Mismatch}
Knowledge distillation~\citep{hinton2015distilling} transfers capability via soft targets and underpins language-model compression~\citep{gu2024minillm,ko2024distillm}.
A key recent advance replaces off-policy supervision with \emph{on-policy} distillation, training the student on its own rollouts under teacher guidance~\citep{agarwal2024onpolicy,agarwal2024gkd,xu2024onpolicy}, which eliminates the student-side distribution mismatch that limits offline self-distillation for reasoning.
On-Policy Self-Distillation (OPSD)~\citep{zhao2026opsd} sharpens this by letting a single model play both roles: the teacher conditions on a complete ground-truth solution as privileged information and provides dense token-level supervision along the student's own on-policy rollouts.
Concurrent work extends the paradigm to diverse feedback formats, continual fine-tuning, reasoning compression, and RL hybrids~\citep{hubotter2026sdpo,shenfeld2026sdft,sang2026opsdc,stein2026gates,ding2026hdpo}, while recent analyses characterise its stability across supervision signals and model scales~\citep{li2026rethinking,chen2026soda}.
Meanwhile, teacher--student mismatch has long been addressed on the student side---via scheduled sampling~\citep{bengio2015scheduled}, DAgger-style imitation~\citep{ross2011dagger}, and importance reweighting~\citep{li2026rethinking,yan2026dasd}---but all such efforts adjust only the student's training distribution and leave the teacher's conditioning unchanged.

In this paper, we observe that the teacher's access to privileged reasoning is treated as a fixed binary choice (full or none) across all prior work, with only the student side being adapted.
To this end, we formulate teacher exposure as a continuous, learnable control variable on top of OPSD, turning it from a fixed default into a training-state-conditioned decision about the teacher's privileged context.

\subsection{Adaptive Distillation Curricula and Learned Control}
Adaptive distillation has so far modulated the student's view of a \emph{fixed} teacher: curriculum learning orders examples by difficulty~\citep{bengio2009curriculum}; dynamic-temperature schedules tie the distillation softmax to sample difficulty~\citep{li2023curriculum}, adversarial signals~\citep{jin2025adversarial}, logit correlations~\citep{matsuyama2025adaptive}, or training state~\citep{islam2025dynamic}; and stronger adaptive teachers further tune their teaching strategy to student progress~\citep{huang2025distplus}.
A separate reinforcement-learning line enhances LLM reasoning via PPO~\citep{schulman2017ppo}, DPO~\citep{rafailov2023direct}, and rule-reward systems such as DeepSeek-R1~\citep{guo2025deepseekr1} and DAPO~\citep{yu2025dapo}; these methods also show that delayed effects often require credit assignment beyond same-step rewards~\citep{xu2025direct,setlur2024rl,tan2026hcapo}.
In this paper, we introduce a different form of adaptation: rather than adjusting the student's view of a fixed teacher, we modulate the teacher's own information level and learn this exposure control via REINFORCE with a discounted learning-progress reward over later student updates rather than same-step loss changes.

\section{Preliminaries}
\label{sec:prelim}

\subsection{On-Policy Self-Distillation}
\label{sec:opsd}

We build upon On-Policy Self-Distillation (OPSD)~\citep{zhao2026opsd}, which instantiates both a teacher and a student policy from a single language model $p_\theta$ by varying the conditioning context. Given a reasoning dataset $\datacal = \{(x_i, \refsol_i)\}_{i=1}^N$, OPSD defines a \textbf{student policy} $\spolicy(\cdot \mid x) \triangleq p_\theta(\cdot \mid x)$, conditioned only on the problem, and a \textbf{teacher policy} $\tpolicy(\cdot \mid x, \refsol) \triangleq p_\theta(\cdot \mid x, \refsol)$, conditioned on the problem and full reference solution. Training samples an on-policy rollout $\rollout \sim \spolicy(\cdot \mid x)$ and minimizes the per-token forward KL between teacher and student conditional distributions along the same rollout:
\begin{equation}
  \loss_{\text{OPSD}}(\theta) = \mathbb{E}_{(x, \refsol) \sim \datacal} \; \mathbb{E}_{\rollout \sim \spolicy(\cdot|x)} \left[ \frac{1}{|\rollout|} \sum_{n=1}^{|\rollout|} \KL\!\left( \tpolicy(\cdot \mid x, \refsol, \rollout_{<n}) \;\|\; \spolicy(\cdot \mid x, \rollout_{<n}) \right) \right],
  \label{eq:opsd}
\end{equation}
Gradients flow only through $\spolicy$; the teacher $\tpolicy$ is treated as a frozen dense target informed by $\refsol$, with pointwise KL clipping used for stability. A critical assumption in Eq.~\eqref{eq:opsd} is that the teacher always conditions on the \emph{complete} reference solution $\refsol$---a default inherited by follow-up methods without justification. We next examine whether this assumption actually yields optimal supervision.

\subsection{A Closer Look at Teacher Exposure}
\label{sec:closer_look}

Although OPSD achieves strong performance, its teacher exposure is fixed at full reveal, and no prior work tests whether complete access to $\refsol$ gives the best supervision throughout training. We therefore formalize teacher exposure as a continuous analytical variable, opening this previously unexamined default to direct empirical study and systematic measurement of supervision quality across $\alpha$.

\paragraph{Teacher exposure as a continuous variable.}
We introduce an \textbf{exposure fraction} $\alpha \in [0, 1]$ controlling how much reference reasoning the teacher sees. Let $\refsol = (\refsol_{\text{reason}}, \refsol_{\text{answer}})$ denote the reasoning trace and the final boxed answer of the reference solution. Given an exposure level $\alpha$ -- interpreted as a fraction of the privileged reasoning prefix -- we construct an exposed reference
\begin{equation}
  \text{truncate}(\refsol, \alpha) = \left( \refsol_{\text{reason}}[1 : \lfloor \alpha \cdot |\refsol_{\text{reason}}| \rfloor], \; \refsol_{\text{answer}} \right),
  \label{eq:truncate}
\end{equation}
where the final answer is always preserved. The exposure-modulated teacher at exposure level $\alpha$ is
\begin{equation}
  \tpolicy^\alpha(\cdot \mid x, \refsol) \triangleq p_\theta\!\left(\cdot \mid x, \;\text{truncate}(\refsol, \alpha)\right).
  \label{eq:teacher_alpha}
\end{equation}
Here $\alpha = 1$ recovers standard OPSD, while $\alpha = 0$ gives only the final answer. We define the expected per-token teacher--student mismatch at exposure level $\alpha$ along the on-policy student rollouts as
\begin{equation}
  M(\alpha) = \mathbb{E}_{(x,\refsol) \sim \datacal} \; \mathbb{E}_{\rollout \sim \spolicy(\cdot|x)} \left[ \frac{1}{|\rollout|} \sum_{n=1}^{|\rollout|} \KL\!\left( \tpolicy^\alpha(\cdot \mid x, \refsol, \rollout_{<n}) \;\|\; \spolicy(\cdot \mid x, \rollout_{<n}) \right) \right].
  \label{eq:mismatch}
\end{equation}
As $\alpha$ increases, the teacher conditions on more privileged reasoning and its predictive distribution $\tpolicy^\alpha$ becomes sharper, concentrating probability on tokens consistent with the reference trace while $\spolicy$ remains unchanged. This widening KL measures supervision that is increasingly informative, but also increasingly difficult for the current student to absorb without an explicit controller adjustment.

\begin{figure}[t]
\centering
\includegraphics[width=\linewidth]{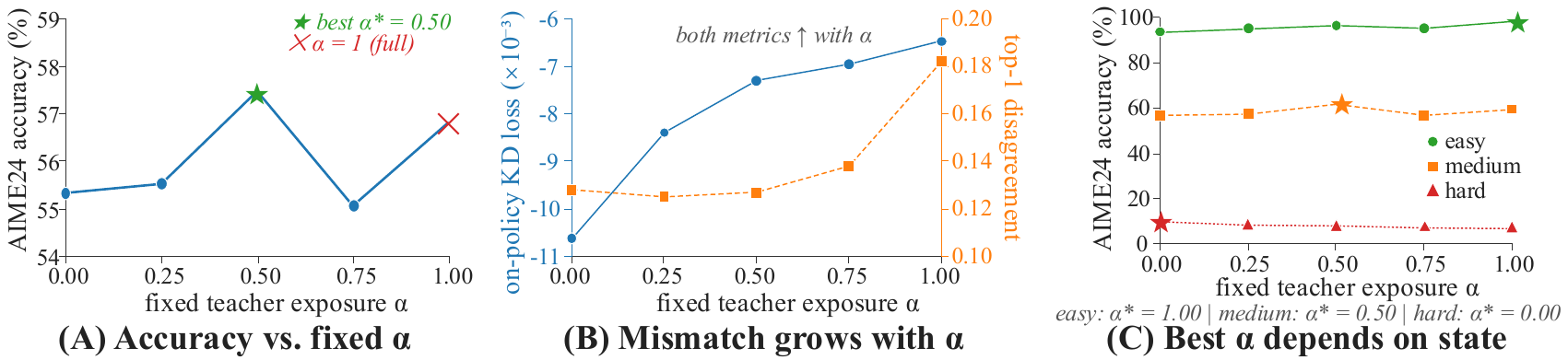}
\caption{Empirical analysis of teacher exposure on AIME\,2024 with Qwen3-1.7B (3 seeds, mean$\pm$s.e.m.).
\textbf{(A)}~Accuracy vs.\ fixed $\alpha$: the best fixed exposure is intermediate ($\alpha^{*}{=}0.5$), not full exposure.
\textbf{(B)}~Both mismatch proxies (on-policy KD loss tail, blue; top-1 disagreement, orange) grow monotonically with $\alpha$.
\textbf{(C)}~Best observed grid exposure varies by difficulty: among $\{0,0.25,0.5,0.75,1.0\}$, easy prefers $1.0$, medium prefers $0.5$, and hard prefers the lowest tested exposure, so no single fixed value serves all learning regimes equally well across difficulty tiers.}
\label{fig:closer_look}
\end{figure}

\paragraph{Empirical verification.}
A natural question is whether full exposure is actually optimal. We sweep $\alpha \in \{0, 0.25, 0.5, 0.75, 1.0\}$ across 3 seeds (Figure~\ref{fig:closer_look}) and observe three patterns. \emph{Suboptimality of Full Exposure} (Figure~\ref{fig:closer_look}A): the best fixed value is intermediate ($\alpha^{*}{=}0.5$), not full exposure, so more privileged information does not automatically yield better supervision. \emph{Monotonic Mismatch Growth} (Figure~\ref{fig:closer_look}B): both on-policy KD loss tail and top-1 disagreement increase with $\alpha$, matching the trend predicted by $M(\alpha)$. \emph{Exposure Depends on Learning Regime} (Figure~\ref{fig:closer_look}C): the best observed grid value differs across easy, medium, and hard samples, with the hard bin preferring the lowest tested exposure. This does not imply that answer-only supervision is universally optimal for hard problems; rather, under this coarse grid it shows that full reasoning exposure can exceed what the current student can absorb.
Thus the issue is not simply that full exposure is ``too much'' in all cases; rather, exposure must match what the student can currently use. This turns teacher exposure from a static prompt-design choice into a training-time control problem.
Importantly, the student-side rollout protocol is unchanged across the sweep. The observed trend is therefore induced by the teacher's privileged context rather than by a different sampling distribution, isolating exposure as the variable that modulates supervision while the rest of the OPSD recipe stays unchanged for fair comparison.

\paragraph{Teacher-side exposure mismatch.}
Taken together, these findings identify a \emph{teacher-side exposure mismatch}: as $\alpha$ grows, teacher targets can drift outside the student's learnable range. This is the supervision-side analogue of the on-policy rollout mismatch that OPSD removes on the student side. The natural response is to treat $\alpha$ as a learnable training-time control variable rather than a fixed default. In the next section, we propose \methodname{} to achieve this.
This distinction also clarifies the scope of the method: we do not change how student rollouts are collected, but only change how much privileged reasoning the teacher may use when scoring those rollouts during on-policy distillation.

\section{Method: \methodname}
\label{sec:method}

\begin{figure}[t]
\centering
\includegraphics[width=\linewidth]{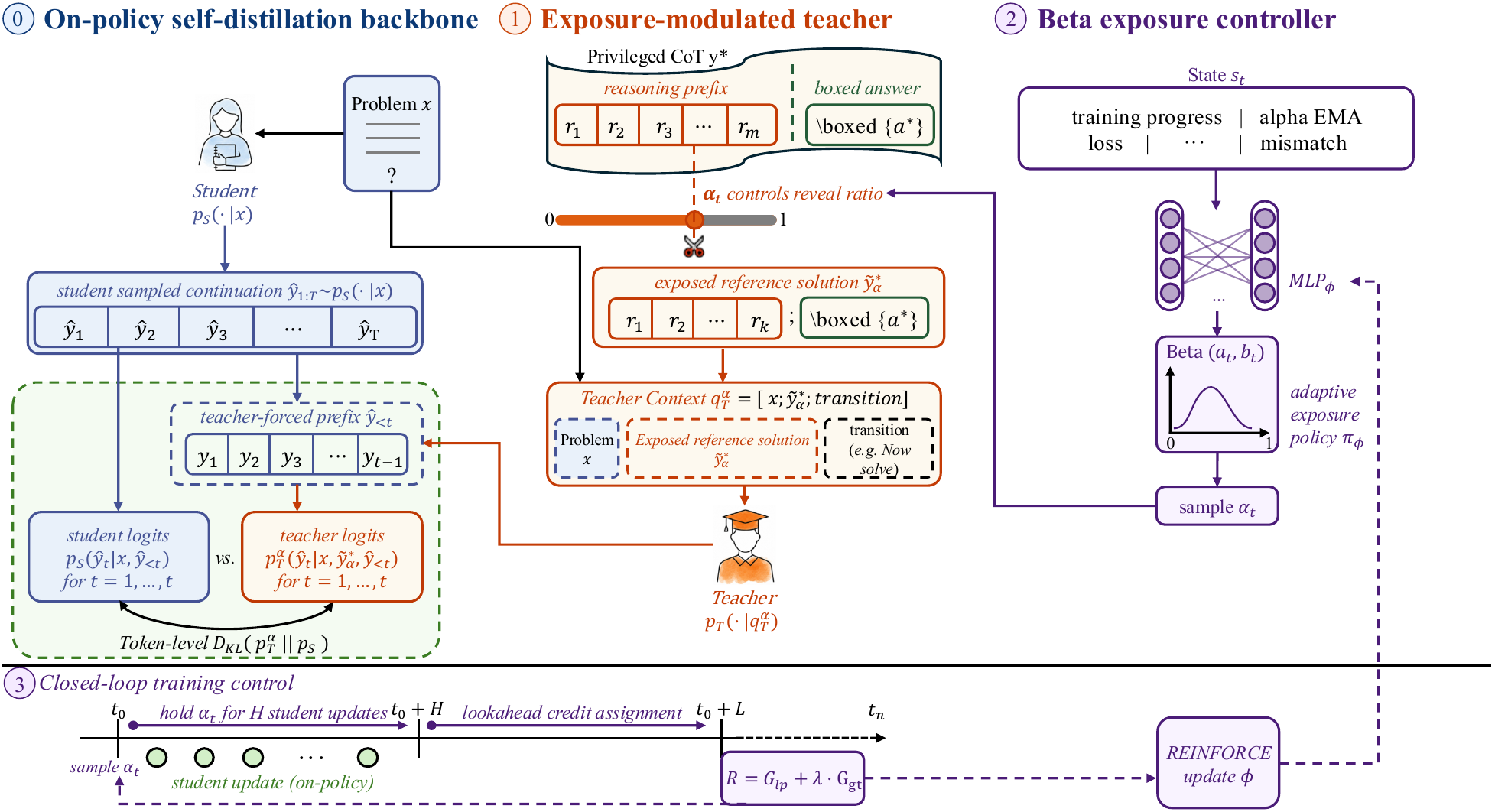}
\caption{Overview of \methodname. The OPSD backbone samples student continuations from the problem-only prompt. Given an exposure action $\alpha_t$, \methodname{} truncates only the reasoning prefix of the privileged solution, preserves the boxed answer, and builds the teacher context from the problem, exposed reference solution, and transition prompt. Teacher and student logits are compared on the same teacher-forced student tokens. A training-state-conditioned Beta controller samples one global $\alpha_t$ for a hold window, then receives a delayed lookahead reward for its REINFORCE policy updates.}
\label{fig:method}
\end{figure}

Section~\ref{sec:closer_look} turns the full-reference teacher in OPSD into three design requirements.
First, full exposure is not reliably optimal, so teacher exposure should be continuous rather than binary.
Second, teacher--student mismatch grows with exposure, so the exposure level should be chosen from training feedback instead of fixed by hand.
Third, the effect of an exposure decision is only visible after subsequent student updates, so the controller needs delayed credit rather than a one-step loss proxy.
\methodname{} implements these requirements while keeping the OPSD student rollout unchanged.
As shown in Figure~\ref{fig:method}, it replaces the full-reference teacher with an exposure-modulated teacher, samples one global exposure for each hold window using a training-state-conditioned Beta controller, and credits that held action through a closed-loop lookahead reward over later student updates in training.

\subsection{Exposure-Modulated Teacher}
\label{sec:exposure_teacher}

We implement the first module by replacing the full reference context in OPSD with an $\alpha_t$-controlled teacher context.
Given a sampled exposure $\alpha_t$, we truncate the reference solution and insert the exposed reference into the teacher prompt used for teacher scoring during token-level distillation:
\begin{equation}
  \begin{aligned}
  \tilde{\refsol}_{\alpha_t}
  &=
  \operatorname{truncate}(\refsol,\alpha_t),
  \\
  q_{\mathrm{T}}^{\alpha_t}(x,\refsol)
  &=
  [\,x;\tilde{\refsol}_{\alpha_t};\tau\,],
  \end{aligned}
  \label{eq:teacher_context}
\end{equation}
where $\tau$ is a fixed transition instruction.
The truncation acts only on the reasoning prefix; the final boxed answer is retained.
Thus $\alpha_t$ controls how much privileged reasoning the teacher sees while keeping the answer constraint available for every exposure level.
This simple prefix operator preserves reasoning order while isolating how much privileged context the teacher uses during teacher scoring.

Training remains on-policy on the student side.
The student samples a continuation $\rollout_{1:T}\sim\spolicy(\cdot\mid x)$ from the problem-only prompt.
We teacher-force the same sampled tokens through two contexts: the student context $[x;\rollout_{<n}]$ and the teacher context $[q_{\mathrm{T}}^{\alpha_t};\rollout_{<n}]$.
The rollout therefore supplies a common scoring prefix, while $\alpha_t$ changes only the teacher's privileged information.
The objective is the OPSD token-level KL with the full-reference teacher replaced by the exposure-modulated teacher:
\begin{equation}
  \loss_{\text{ATESD}}(\theta;\alpha_t)
  =
  \mathbb{E}_{(x,\refsol)\sim\datacal,\;\rollout\sim\spolicy(\cdot\mid x)}\!\left[
  \frac{1}{|\rollout|}
  \sum_{n=1}^{|\rollout|}
  \KL\!\left(
  \tpolicy^{\alpha_t}(\cdot \mid x,\refsol,\rollout_{<n})
  \;\|\;
  \spolicy(\cdot \mid x,\rollout_{<n})
  \right)
  \right].
  \label{eq:atesd_loss}
\end{equation}
Gradients flow only through the student.
Low exposure weakens the teacher's reasoning context without corrupting the answer, while high exposure recovers the standard full-reference teacher.

\subsection{Beta Exposure Controller}
\label{sec:controller}

The controller chooses an information intensity rather than a discrete curriculum label.
We parameterize it as a training-state-conditioned Beta policy $\pi_\phi(\alpha\mid s_t)$.
The state $s_t$ summarizes global training progress, recent exposure, loss and mismatch EMAs, a probe-NLL EMA, and batch-aggregated student self-confidence.
A lightweight MLP then maps this compact state to the concentration parameters defining the continuous exposure policy used throughout the held action window:
\begin{equation}
  (a_t,b_t)=1+\operatorname{softplus}(f_\phi(s_t)), \qquad
  \alpha_t \sim \Beta(a_t,b_t), \qquad
  \alpha_t \leftarrow \operatorname{clip}(\alpha_t,\alpha_{\min},\alpha_{\max}).
  \label{eq:beta_policy}
\end{equation}
The constraint $a_t,b_t>1$ keeps the Beta distribution unimodal: its mean represents the preferred exposure level, and its concentration represents confidence.
After sampling an action, \methodname{} holds a single $\alpha_t$ fixed for all samples over the next $H$ student updates before resampling.
One exposure decision therefore controls a short global episode, which is credited by later loss changes and teacher-grounded scores over the entire held window rather than a single minibatch.

\subsection{Closed-Loop Training Control}
\label{sec:reward}
\label{sec:training}

We train the controller with a closed-loop schedule because exposure decisions have delayed effects.
A high-exposure action may help future learning even if its immediate loss drop is small, while a low-exposure action may look safe but provide little pressure.
For an action sampled at step $t_0$, \methodname{} holds $\alpha_t$ for $H$ student updates and scores it after an $L$-step lookahead window.
The reward combines discounted learning progress with a teacher-grounded credit score for the held action:
\begin{equation}
  \begin{aligned}
  G_{\text{lp}}(t_0)
  &=
  \sum_{i=1}^{L}\gamma^{i-1}
  \max\!\left(0,\ell_{t_0+i-1}-\ell_{t_0+i}\right),\\
  G_{\text{gt}}(t_0)
  &=
  \frac{\sum_{i=1}^{L}\gamma^{i-1} g_{t_0+i}}
       {\sum_{i=1}^{L}\gamma^{i-1}},\\
  R(t_0)
  &=
  G_{\text{lp}}(t_0)+\lambda_{\text{gt}}G_{\text{gt}}(t_0).
  \end{aligned}
  \label{eq:reward}
\end{equation}
Here $\ell_t$ is the distillation loss after step $t$, and $g_t$ is the average log-probability assigned by the exposure-modulated teacher to verified reference tokens.
The first term rewards realized positive student improvement; the second keeps high-reward actions tied to a teacher that still predicts the ground-truth solution.
Clipping stabilizes the reward scale; the centered advantage in Eq.~\eqref{eq:advantage} still gives below-average held actions negative policy updates.
Teacher--student mismatch is used as controller state and diagnostic signal, not as a direct reward penalty in the main objective, because such a penalty would prefer low exposure simply for mechanically reducing KL against the student.

The student is updated every step using Eq.~\eqref{eq:atesd_loss}, while the controller is updated only after held actions complete their lookahead windows.
For a batch of completed decisions $\{(\alpha_j,s_j,R_j)\}_{j=1}^{B}$, we center and normalize rewards before applying REINFORCE to update the held-action Beta exposure policy:
\begin{equation}
  A_j = \frac{R_j-\bar{R}}{\operatorname{Std}(R)+\epsilon},
  \qquad
  \loss_{\text{ctrl}}
  =
  -\frac{1}{B}\sum_{j=1}^{B}A_j\log\pi_\phi(\alpha_j\mid s_j)
  + c_t\max\!\left(0,\mathcal{H}[\pi_\phi]-\mathcal{H}_{\text{target}}\right)^2 .
  \label{eq:advantage}
\end{equation}
The entropy term only caps persistent over-exploration; it still allows the policy to concentrate when delayed feedback consistently favors a narrower exposure region as training enters a stable regime.

\section{Experiments}
\label{sec:experiments}

\begin{table}[t]
  \caption{Main results on competition-level mathematical reasoning benchmarks. We follow the OPSD reporting protocol and report Average@12 accuracy (\%) under the Qwen3 sampling configuration. Baseline numbers are from \citet{zhao2026opsd}; \methodname{} is evaluated with the same within-100-step checkpoint selection convention. Best results are in \textbf{bold}; second-best results are \underline{underlined}.}
  \label{tab:main}
  \centering
  \small
  \begin{tabular}{l ccc c}
    \toprule
    \textbf{Method} & \textbf{AIME\,24} & \textbf{AIME\,25} & \textbf{HMMT\,25} & \textbf{Average} \\
    \midrule
    \multicolumn{5}{l}{\textit{Qwen3-1.7B}} \\
    \quad Base (Instruct)~\citep{team2025qwen3} & 51.5 & 36.7 & 23.1 & 37.1 \\
    \quad + SFT~\citep{ouyang2022instructgpt}   & 48.4 & 36.3 & 22.7 & 35.8 \\
    \quad + GRPO~\citep{shao2024deepseekmath}   & 51.1 & 38.3 & 23.7 & 37.7 \\
    \quad + OPSD~\citep{zhao2026opsd}            & \underline{57.2} & \underline{43.9} & \textbf{29.2} & \underline{43.4} \\
    \quad + Ours                      & \textbf{59.17}\seedstd{0.8} & \textbf{44.72}\seedstd{0.28} & \underline{29.17}\seedstd{1.37} & \textbf{44.35}\seedstd{0.23} \\
    \midrule
    \multicolumn{5}{l}{\textit{Qwen3-4B}} \\
    \quad Base (Instruct)~\citep{team2025qwen3} & 74.9 & 66.4 & 42.2 & 61.2 \\
    \quad + SFT~\citep{ouyang2022instructgpt}   & 70.2 & 62.3 & 43.4 & 58.6 \\
    \quad + GRPO~\citep{shao2024deepseekmath}   & 75.6 & 68.1 & 44.4 & 62.7 \\
    \quad + OPSD~\citep{zhao2026opsd}            & \underline{76.4} & \underline{68.3} & \underline{46.1} & \underline{63.6} \\
    \quad + Ours                      & \textbf{78.06}\seedstd{0.43} & \textbf{71.39}\seedstd{0.48} & \textbf{47.50}\seedstd{0.89} & \textbf{65.65}\seedstd{1.04} \\
    \midrule
    \multicolumn{5}{l}{\textit{Qwen3-8B}} \\
    \quad Base (Instruct)~\citep{team2025qwen3} & 75.8 & 65.6 & 43.9 & 61.8 \\
    \quad + SFT~\citep{ouyang2022instructgpt}   & 72.3 & 64.2 & 42.9 & 59.8 \\
    \quad + GRPO~\citep{shao2024deepseekmath}   & 76.4 & 68.9 & \underline{46.7} & 64.0 \\
    \quad + OPSD~\citep{zhao2026opsd}            & \underline{77.8} & \underline{70.8} & 45.8 & \underline{64.8} \\
    \quad + Ours                      & \textbf{80.56}\seedstd{0.70} & \textbf{72.50}\seedstd{0.32} & \textbf{48.33}\seedstd{0.28} & \textbf{67.13}\seedstd{0.32} \\
    \bottomrule
  \end{tabular}
\end{table}

We evaluate \methodname{} on competition-level mathematical reasoning.
Section~\ref{sec:closer_look} already answers the diagnostic questions: full exposure is not reliably optimal, and mismatch increases as the teacher sees more privileged reasoning.
The experiments below ask whether exposure learning improves OPSD and whether the ablations support the exposure-control mechanism under the same setup and budget.

\subsection{Experimental Settings}
\label{sec:setup}

\paragraph{Setup.}
We validate \methodname{} on instruct-tuned Qwen3-1.7B, Qwen3-4B, and Qwen3-8B models~\citep{team2025qwen3}.
Following OPSD~\citep{zhao2026opsd}, all post-training methods use the OpenThoughts mathematical reasoning corpus~\citep{guha2025openthoughts} and the same $100$-step on-policy distillation budget.
\methodname{} keeps the OPSD student rollout, optimizer, LoRA training recipe, and problem-only prompting protocol unchanged; it only replaces the full-reference teacher context with a learned exposure policy.
\ifdefined\arxivbodyonly
\else
Detailed optimizer, LoRA, rollout-length, sampling, and hardware settings are left to Appendix~\ref{app:hyperparams} for reproducibility.
\fi

\paragraph{Metrics and baselines.}
We evaluate on AIME\,2024, AIME\,2025, and HMMT\,2025 using Average@12, the mean accuracy over $12$ sampled completions under the OPSD sampling protocol.
Following the OPSD within-budget convention, saved checkpoints inside the $100$-step training budget are evaluated and the best Average@12 score is reported for each benchmark.
Table~\ref{tab:main} compares the instruct base model, SFT~\citep{ouyang2022instructgpt}, GRPO~\citep{shao2024deepseekmath}, and OPSD~\citep{zhao2026opsd}.
The baseline rows are taken from \citet{zhao2026opsd} to match the original reporting convention, model family, datasets, sampling protocol, and checkpoint-selection rule used there for all baseline methods reported in Table~\ref{tab:main} for fairness.

\paragraph{Controller configuration.}
The exposure controller is intentionally small: a 2-layer MLP maps six training-state statistics to a Beta distribution over $\alpha\in[0,1]$.
All main runs use the same lookahead horizon $L=20$ for delayed credit assignment.
\ifdefined\arxivbodyonly
\else
Scale-specific hold windows, discount factors, reward weights, and other implementation details for the reported runs are listed in Appendix~\ref{app:hyperparams}.
\fi

\subsection{Results and Discussion}
\label{sec:main_results}

\paragraph{Adaptive exposure improves OPSD across model scales.}
Table~\ref{tab:main} gives the primary comparison across three model scales and three benchmarks.
\methodname{} achieves the best average performance at every scale under the OPSD reporting protocol.
It improves OPSD by $+0.95$, $+2.05$, and $+2.33$ Average@12 points on Qwen3-1.7B, Qwen3-4B, and Qwen3-8B.
At 1.7B, it improves both AIME benchmarks and matches OPSD on HMMT within $0.03$ points.
At 4B and 8B, it improves OPSD on all three datasets.
The strongest 4B run reaches $65.65$ Average@12, a $2.05$-point gain over OPSD and a $2.95$-point gain over GRPO.
The 8B run further raises the average to $67.13$ on the same benchmark suite, so the trend is not confined to one larger model scale or benchmark subset in Table~\ref{tab:main}.
The gain is larger at 4B and 8B than at 1.7B.
This is consistent with exposure control becoming more valuable when the student has enough capacity to exploit privileged teacher context but still needs that context to be regulated.
At the smallest scale, the usable headroom between the exposed teacher and the student is more limited, so the improvement remains positive but more modest.
Together with the controlled exposure analysis in Section~\ref{sec:closer_look}, this scale pattern confirms that the learned exposure mechanism improves end-task accuracy across model scales rather than merely reflecting the diagnostic sweep that originally motivated the method and the controller design it uses.

\subsection{Ablation Study}
\label{sec:ablation}

\begin{figure}[H]
  \centering
  \includegraphics[width=\linewidth]{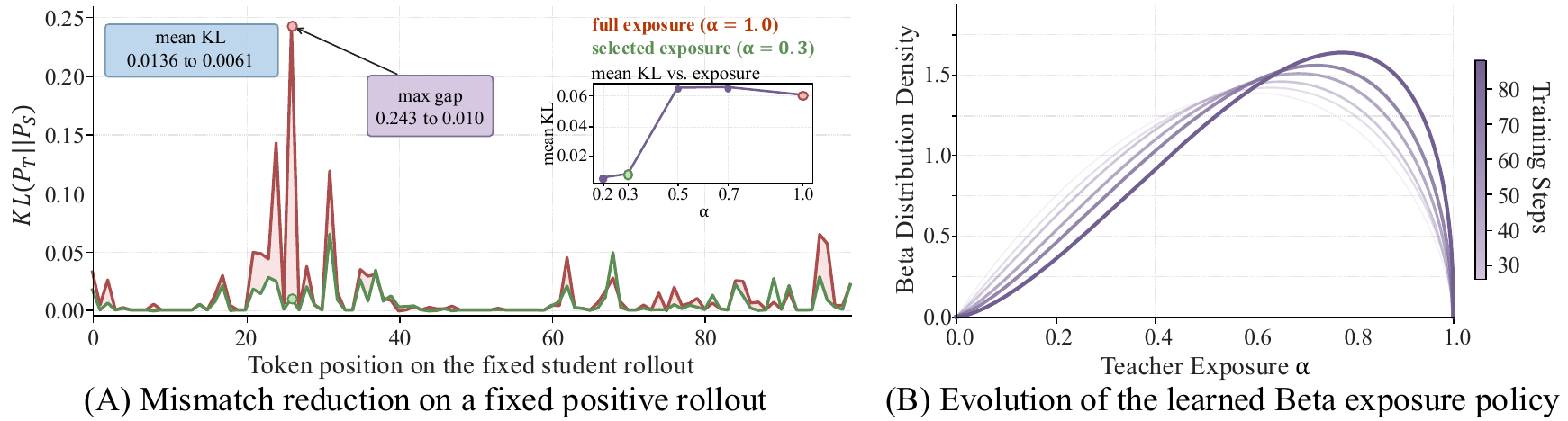}
  \caption{Mechanism ablations for exposure control. \textbf{(A)}~Keeping the problem, student rollout, and scoring positions fixed, reducing teacher exposure lowers the token-level teacher--student KL spikes on a positive trajectory. \textbf{(B)}~The learned Beta policy evolves from broad exploration toward a structured exposure distribution rather than collapsing to either no reference or full exposure.}
  \label{fig:ablation_mixed}
\end{figure}

\paragraph{Can exposure control reduce positive-trajectory supervision mismatch?}
Figure~\ref{fig:ablation_mixed}~(A) fixes the problem, sampled student rollout, and scoring positions.
It changes only the teacher context, from full exposure ($\alpha=1.0$) to the reproduced \methodname{} exposure ($\alpha=0.3$).
On this positive trajectory, exposure control reduces unnecessary supervision mismatch: mean KL drops from $0.0136$ to $0.0061$, max KL drops from $0.2432$ to $0.0645$, and the largest spike at position $26$ drops from $0.2432$ to $0.0098$.
This is the setting where self-distillation should be easiest to exploit.
The student is already producing a useful continuation, so a large KL spike reflects avoidable teacher-context mismatch rather than a need for stronger correction.
The result therefore does not claim that lower mismatch is always better.
It shows that full privileged exposure can over-constrain positive trajectories, while adaptive exposure keeps the same useful trajectory easier to distill under an identical replay protocol and fixed token-scoring positions during the controlled teacher-scoring replay for this analysis.
Because the sampled student continuation is unchanged, this reduction cannot be attributed to an easier trajectory or a different rollout distribution; it comes solely from changing how much privileged reasoning the teacher conditions on during teacher scoring in on-policy distillation.

\paragraph{What training-time exposure policy does \methodname{} learn?}
Figure~\ref{fig:ablation_mixed}~(B) visualizes the Beta exposure distribution over training.
It starts broad and then concentrates away from both no-reference and full-exposure extremes, showing that the controller learns a training-state-level exposure policy rather than using a fixed or uncontrolled choice throughout the held-window control loop.
This interior concentration is important for the paper's main claim: if exposure control merely rediscovered a trivial always-low or always-full rule, the learned policy would collapse toward one boundary and reduce to another fixed heuristic.
Instead, the mass remains in a usable middle regime, consistent with Section~\ref{sec:closer_look} that different learning regimes favor different exposure levels and that successful adaptation should stay within the exposure range rather than anneal to a single universal extreme throughout training.
The learned policy therefore behaves as a continuously adjusted exposure controller rather than a post-hoc selection among a few fixed exposure settings.

\begin{table}[t]
\centering
\small
\caption{Controller ablations on AIME\,2024 Average@12 (\%). \textbf{(A)}~Delayed credit for learning exposure; \textbf{(B)}~learned exposure versus fixed or uncontrolled alternatives. Both subtables use the same evaluation setting, isolating controller design choices rather than data or checkpoint-budget changes.}
\label{tab:controller_ablation}
\begin{minipage}[t]{0.48\linewidth}
\centering
\textbf{(A) Credit signal}\par
\begin{tabular*}{\linewidth}{@{\extracolsep{\fill}}lc@{}}
\toprule
\textbf{Controller signal} & \textbf{Avg@12} \\
\midrule
Immediate one-step & 52.22 \\
Short-horizon delayed & 56.11 \\
Discounted lookahead & 58.06 \\
Full delayed reward (ATESD) & \textbf{59.17} \\
\bottomrule
\end{tabular*}
\end{minipage}

\hfill
\begin{minipage}[t]{0.48\linewidth}
\centering
\textbf{(B) Exposure policy}\par
\begin{tabular*}{\linewidth}{@{\extracolsep{\fill}}lc@{}}
\toprule
\textbf{Policy} & \textbf{Avg@12} \\
\midrule
OPSD full exposure $\alpha=1.0$ & 57.20 \\
Best fixed $\alpha=0.5$ & 57.44 \\
Stochastic exposure & 54.94 \\
Learned policy & \textbf{59.17} \\
\bottomrule
\end{tabular*}
\end{minipage}

\end{table}
\FloatBarrier

\paragraph{Is delayed credit necessary for learning exposure?}
Table~\ref{tab:controller_ablation}~(A) gradually adds the credit-assignment structure used by \methodname{}.
Immediate one-step feedback reaches $52.22$ Average@12, while introducing delayed credit already raises the score to $56.11$.
Replacing the short delayed signal with discounted lookahead further improves performance to $58.06$, and the full delayed reward used by \methodname{} reaches $59.17$ after adding the teacher-grounded score.
This pattern matches the training dynamics of on-policy distillation.
The sampled exposure changes the teacher target used for the current update, but its consequence is visible only after later student optimization steps and refreshed rollouts.
The ablation therefore supports delayed credit as an enabling mechanism for learning exposure, not as an incidental implementation detail or a noisy same-minibatch reward alone.

\paragraph{Does learned exposure outperform fixed or uncontrolled exposure?}
Table~\ref{tab:controller_ablation}~(B) compares fixed exposure, uncontrolled stochastic exposure, and the learned policy.
OPSD full exposure reaches $57.20$, the best fixed exposure reaches $57.44$, and uncontrolled stochastic exposure falls to $54.94$.
The learned policy reaches $59.17$, which rules out two weaker explanations.
It is not merely avoiding full exposure with a manually tuned constant.
It is also not merely injecting stochasticity into the teacher context.
The useful signal is feedback-driven adaptation of $\alpha$ to the training state, consistent with the learned Beta-policy evolution in Figure~\ref{fig:ablation_mixed}.
This differs from a post-hoc fixed choice selected after observing the fixed-exposure sweep or final benchmark outcome.
It is the training-state-level counterpart of the coarse difficulty-bin diagnosis in Figure~\ref{fig:closer_look}C.
The current controller does not choose a separate $\alpha$ for each example, but it learns a global exposure distribution that changes with the training state instead of committing to one exposure for all regimes throughout training.

\section{Conclusion}
\label{sec:conclusion}

We have identified teacher-side exposure mismatch as an overlooked bottleneck in on-policy self-distillation for LLM reasoning.
Through systematic experiments, we established that full teacher exposure is suboptimal and that the teacher--student distribution mismatch grows monotonically with exposure level.
To address this, we proposed \methodname, which learns to control teacher exposure via a training-state-conditioned Beta controller optimized with a discounted learning-progress reward and a hold-and-lookahead scheme for delayed credit assignment across student updates during training.

Experiments on AIME\,2024, AIME\,2025, and HMMT\,2025 across Qwen3-\{1.7B, 4B, 8B\} demonstrate that \methodname consistently improves over the OPSD baseline under the same evaluation protocol.
Ablations further show three complementary effects: exposure control reduces positive-trajectory mismatch, delayed credit is necessary for learning exposure, and feedback-driven exposure selection outperforms fixed or uncontrolled choices under the same student rollout protocol during training.

\paragraph{Limitations and future directions.}
The current controller operates at the \emph{global} level, selecting a single $\alpha$ for all samples within a hold period.
A natural extension is \emph{per-sample} or difficulty-aware exposure control, where $\alpha$ is conditioned on problem difficulty or the student's confidence.
Our coarse difficulty-bin analysis motivates this direction, while the present method deliberately studies the simpler training-state-level controller first.
Additionally, our discounted learning-progress reward relies on a fixed lookahead window.
Counterfactual or model-based reward estimation could further improve credit assignment.
Finally, validating \methodname on larger model scales, code generation, and scientific reasoning remains important for testing whether exposure control extends beyond math contests and the benchmark suite studied here to broader reasoning domains in future studies.

\bibliographystyle{plainnat}
\bibliography{references}

\end{document}